\documentclass{article}
\pdfoutput=1
\usepackage{microtype}
\usepackage{graphicx}
\usepackage{subfigure}
\usepackage{booktabs} 
\usepackage{amsmath}
\usepackage{multirow}
\usepackage{float}
\usepackage{amssymb,amsmath,epsfig}
\usepackage{scalerel}
\usepackage{microtype}
\usepackage{graphicx}
\usepackage{multirow}
\usepackage{booktabs}
\usepackage{amssymb,amsmath,graphicx}
\usepackage{caption}

\usepackage[export]{adjustbox}
\usepackage{moredefs}
\usepackage{color}\restylefloat{table}

\usepackage{hyperref}

\usepackage{enumitem}

\usepackage[accepted]{sysml2019}

\sysmltitlerunning{Optimizing Speech Recognition for the Edge}

\begin{document}

\twocolumn[
\sysmltitle{Optimizing Speech Recognition for the Edge}

\sysmlsetsymbol{equal}{*}

\begin{sysmlauthorlist}
\sysmlauthor{Yuan Shangguan*}{ed}
\sysmlauthor{Jian Li*}{ed}
\sysmlauthor{Qiao Liang*}{ed}
\sysmlauthor{Raziel Alvarez}{ed}
\sysmlauthor{Ian McGraw}{ed}
\end{sysmlauthorlist}
\sysmlaffiliation{ed}{Google, Inc., USA}
\sysmlcorrespondingauthor{Yuan Shangguan}{yuansg@google.com}
\sysmlcorrespondingauthor{Jian Li}{jianlijianli@google.com}
\sysmlcorrespondingauthor{Qiao Liang}{wildstone@google.com}

\sysmlkeywords{Machine Learning, Speech Recognition, End-to-End, Encoder-Decoder, SysML}

\vskip 0.3in

\begin{abstract}
While most deployed speech recognition systems today still run on
servers, we are in the midst of a transition towards deployments on edge devices.
This leap to the edge is powered by the progression from traditional speech
recognition pipelines to end-to-end (E2E) neural architectures, and the parallel
development of more efficient neural network topologies and optimization
techniques. Thus, we are now able to create highly accurate speech recognizers
that are both small and fast enough to execute on typical mobile devices. In
this paper, we begin with a baseline RNN-Transducer architecture comprised of
Long Short-Term Memory (LSTM) layers. We then experiment with a variety of
more computationally efficient layer types, as well as apply optimization
techniques like neural connection pruning and parameter quantization to
construct a small, high quality, on-device speech recognizer that is an order
of magnitude smaller than the baseline system without any optimizations.

\end{abstract}
]



\printAffiliationsAndNotice{\sysmlEqualContribution} 

\setlength{\abovedisplayskip}{2pt}
\setlength{\belowdisplayskip}{2pt}

\section{Introduction}
\label{intro}
Whether for image processing~\cite{gokhale2014240} or
for speech applications~\cite{chen2014speech}, neural networks have
been finding their way onto edge devices for the better part of a
decade now. It stands to reason then that the search for ways to make
these networks smaller and faster has become increasingly urgent. The
three predominant ways of doing so are through
quantization~\cite{Alvarez16, jacob17}, sparsity~\cite{lecun1990}, and
architecture variation~\cite{greff2016lstm} or some combination
thereof~\cite{han2016dsd}. This work explores all three to accomplish the
goal of creating an all-neural speech recognizer that runs on-device
in real time.

We begin by examining neural network pruning. Early approaches
explored gradient-based pruning~\cite{lecun1990,hassibi1994optimal}, as well as
magnitude-based pruning~\cite{yu2012exploiting,han2015,zhu2018,frankle2019lt}
from a dense network already trained to reasonable accuracy.
Others have explicitly explored the possibility of obtaining an optimal pruned network
without needing to pre-train a dense network~\cite{frankle2019lt,frankle2019lt2,liu2019,lee2019}.

In this paper, we develop an automated gradual pruning algorithm
to obtain pruned speech recognition models with
fewer parameters and minimal accuracy loss based on work by~\cite{zhu2018}.

After pruning, we examine alternative recurrent neural network (RNN)
layer architectures. The baseline is a fairly standard long short-term
memory (LSTM) architecture first
proposed by Hochreiter and Schmidhuber \cite{hochreiter97}, further
enhanced by a forget gate \cite{gers00}. The two primary alternatives we
explore in this work are the Simple
Recurrent Unit (SRU)~\cite{lei2018simple} and the Coupled Input-Forget
Gate (CIFG)~\cite{greff2016lstm}. Both are less complex conceptually and
computationally than the LSTM.

Finally we look at quantization. Numerous methodologies have been developed to
convert floating point weights into low-bit representations
~\cite{mellempudi2017mixed,zhou2017incremental,courbariaux2016binarized}.
In this paper, we explore low-bit computations either with an efficient mix of 8
and 16-bit integer, or a mix of 8-bit integer and 32 bit
float precision (a.k.a. the \emph{hybrid} approach).

Prior works existed in exploring each of these subtopics, but few have studied the
optimal combination of these techniques to run production End-to-End speech recognition
systems on edge devices. We explain our techniques in
sections~\ref{pruning}, \ref{archs},
and \ref{quant}. Section~\ref{Experiments} delves into experiments on a state-of-the-art
end-to-end Recurrent Neural Network Transducer (RNN-T) speech recognizer.
Section~\ref{conclusions} wraps up with interpretations we derived from our results.

\section{Pruning}
\label{pruning}
Neural networks are often over-parameterized; they occupy expensive computational
resources and create unnecessarily large memory footprints. This motivates network
pruning~\cite{lecun1990, han2015}, through which sparsity is introduced to reduce
model size while maintaining the quality of the original model.

In this work, we adopt the pruning method proposed by Zhu \& Gupta. We increase the sparsity of
weight matrices from an initial value to a final value in a polynomial fashion~\cite{zhu2018}.

Unlike most iterative pruning methods in the literature that zero out the weight matrices
incrementally, our pruned weight elements can be recovered at a later training stage~\cite{zhu2018}.
It is achieved by keeping the values of pruned weights instead
of setting them to zero even though they do not contribute to forward propagation and are
not back propagated. When the mask is updated later, a pruned weight can be recovered if
its retained value is bigger than some un-pruned weights. In our experiments,
we continue to update the mask even after sparsity reaches the target value so that weights pruned
early due to bad initialization can be recovered.

To reduce memory usage and compute more efficiently. We develop a variation of
the Block Compressed Sparse Row (BCSR)~\cite{saad2003} structure to store models.
A stored model contains an array with non-zero blocks in row-major order and
a \emph{ledger} array which contains the number of non-zero blocks of each block
row followed by block column indices of non-zero blocks.

\section{Efficient RNN Variants}
\label{archs}
The long short-term memory (LSTM) cell topology has made
its way into many speech applications: acoustic
models~\cite{sak2014long, mcgraw16}, language models~\cite{jozefowicz16},
and end-to-end models~\cite{chan2016listen, jaitly2016online, chiu2018state, he2019streaming}.
Our baseline LSTM cell is built from the original topology by Hochreiter and
Schmidhuber~\cite{hochreiter97}, further enhanced with a forget gate~\cite{gers00}.

In our LSTM cell implementation (equations~\ref{i1} to~\ref{h1}),
$x^t$ is the input at time $t$, $W$ and $R$ the input
and recurrent weight matrices respectively, $b$ the bias vectors,
$\sigma$ the sigmoid function, $z$, $i$, $f$, $o$ and
$c$ the LSTM-block input, input gate, forget gate, output gate and cell activation vectors,
$h$ the cell output, $\odot$ the element-wise product of the vectors,
and $g$ the activation function, generally $tanh$. We also adopt the
output projection $W_{proj}$ proposed by~\cite{sak2014long} to reduce the cell size.

Layer normalization is added to the input block $z^t$ and gates
($i^t$, $f^t$, and $o^t$). Layer normalization helps to
stabilize the hidden layer dynamics and speeds up model
convergence~\cite{ba2016layer}.

\begin{eqnarray}
i^t = \sigma (W_i x^t + R_i h^{t-1} + b_i)  \label{i1} \\
f^t = \sigma (W_f x^t + R_f h^{t-1} + b_f)  \label{f1} \\
z^t = g(W_z x^t + R_z h^{t-1} + b_z)    \label{z1} \\
c^t = i^t \odot z^t + f^t \odot c^{t-1}  \label{c1} \\
o^t = \sigma(W_o x^t  + R_o h^{t-1} + b_o)  \label{o1} \\
m^t = o^t \odot g(c^t)   \label{m1} \\
h^t = W_{proj} m^t \label{h1}
\end{eqnarray}

\subsection{CIFG-LSTM}
\label{cifg}
The CIFG-LSTM variant is an LSTM modification~\cite{greff2016lstm}.
It is similar to that proposed in the gated recurrent unit (GRU)~\cite{ChoMGBSB14}.
CIFG simplifies the LSTM cell topology by having a single set of
weights (i.e. a single gate) controlling both
the amount of information added to and removed from the cell state.
Multiple work has demonstrated the importance of
the forget gate~\cite{jozefowicz15}, and that removing input gate does not
hurt LSTM cell performances~\cite{greff2016lstm,van2018unreasonable}.

In CIFG, equation~\ref{i1} is replaced with
equation~\ref{i2}. As a result
of this coupling, CIFG has 25\% fewer parameters compared to the baseline LSTM topology.
\begin{equation}
i^t = 1-f^t  \label{i2}
\end{equation}

\subsection{Simple Recurrent Unit (SRU)}
\label{sru}
The SRU cell topology emerges in recent works as an alternative tool for speech and
language processing tasks. Lei et al. show that SRU cells are not only
highly parallelizable in model inferencing but also adequately expressive in capturing the
recurrent statistical patterns in the input text-based data~\cite{lei2018simple}.
Park et al. successfully built an on-device acoustic model by combining
one layer of SRU cells with one layer of depth-wise 1D Convolutional Neural
Network~\cite{park2018fully}.

We implemented a version of SRU in equations~\ref{f2} to~\ref{yt}.
Output projection (equation~\ref{yt}) in the SRU topology reduces
the output dimension of cells. Each layer of SRU also contains layer normalization
on the forget gate $f^t$, reset gate $r^t$, and the cell value $c^t$.

\begin{eqnarray}
f^t = \sigma(W_f x^t  + b_f) \label{f2} \\
r^t = \sigma(W_r x^t  + b_r) \label{r2} \\
c^t = f^t \odot c^{t-1}  + (1-f^t)\odot (W_{x_1} x^t  + b_{x_1})\label{c2} \\
m^t = r^t \odot g(c^t)  + (1-r^t)\odot  (W_{x_2} x^t  + b_{x_2}) \label{h2} \\
h^t = W_{proj}m^t \label{yt}
\end{eqnarray}

SRU and CIFG topologies have not yet been explored in the context of
end-to-end speech recognition models. In this paper, we provide empirical comparisons of the
efficiency, memory and computational latency of these cell topologies.

\section{Quantization}
\label{quant}
Quantization reduces the model's memory consumption (on disk and on RAM), and
speeds up the model’s execution to meet the on-device real-time requirements.
It transforms the model's computation graph such that it
can be (entirely or partially) represented and executed in a lower precision.
In this work, \emph{quantization} refers to the affine
transformation that linearly maps one or more dimensions of a tensor
from a higher to a lower bit representation.

We explored two quantization schemes: \emph{Hybrid quantization} and  \emph{Integer quantization}.
Both schemes have TensorFlow tooling and execution support~\cite{tfmot, tflite},
and both schemes are in the form of \emph{post-training} quantization.

\subsection{Hybrid Quantization}
\label{hybrid-quant}
In the hybrid approach, we operate matrix multiplications in 8-bit
integers precision while representing the matrix multiplication products
in floating point precision for the activation functions~\cite{mcgraw16, Alvarez16, he2019streaming}.
It performs on-the-fly quantization of dynamic values (e.g. activations).
One of the advantages of hybrid quantization is that it can be performed
entirely as a single pass transformation over the graph: no pre-computation is needed.

\subsection{Integer Quantization}
\label{int8-quant}
Integer quantization restricts the computational graph to operate with integer
precision. It requires that the dynamic
tensors be quantized with a scale pre-computed from dev data statistics.
The statistics for all dynamic tensors are collected by running inferences
on a floating point version of the model, logging the dynamic ranges of
each tensor. Integer-only quantization is not only widely supported across
various hardware but also more efficient because
they are faster and consumes less power. The use of
pre-computed scales means there is no overhead re-computing scales with every
inference.

The main challenge of integer quantization for LSTM layers is designing
execution strategy for the operations in the computation graph to minimize
quantization loss. Due to the stateful nature of
LSTM, errors can easily accumulate within the cell.
We follow two principles for
LSTM integer quantization: 1) Matrix related operations,
such as matrix-matrix multiplication are in 8-bit; 2) Vector related operations,
such as element wise sigmoid, are in a mixture of 8-bit and 16-bit. Some
techniques also help to make the computational graph more quantization-friendly.
Layer normalization, for example, improves accuracy for integer only calculation.
It is likely that layer
normalization guards the model against scale shifts caused by quantizing gate
matrix multiplications, which is the primary source of accuracy degradation for
LSTM cell quantization.

During inference, the layer normalization applied to a vector $x_i$ results
in a vector $x_i^{'}$ with zero mean and standard deviation of 1
(equation~\ref{ln_mean} and \ref{ln_norm}). Assuming a
normal distribution, 99.7\% of values in $x_i^{'}$ is confined between $[-3.0, 3.0]$.
This means that the integer representation of $x_i^{'}$ is also limited,
$q_i^{'}\in[-3.0, 3.0]$, which is only 7 values or 2.8 bit. This limit
causes possible accuracy degradation in the model.

\begin{eqnarray}
\bar{x} = \frac{\sum_{i = 1}^{n}(x_i)}{n} \label{ln_mean} \\
x_i^{'} = \frac{(x_i - \bar{x})}{\sqrt{\frac{\sum_{i=1}^{n}(x_i^2 - \bar{x}_i^2)}{n}}} \label{ln_norm}
\end{eqnarray}

We resolved the restriction by adding a scaling factor for
$q_i^{'}$ in the computational graph. With $x_i^{'} = q_i^{'} s^{'}$,
$q_i^{'}$ can now be expressed as
$q_i^{'} = \frac{x_i^{'}}{s^{'}} = \frac{(q_i - \bar{q})}{\sqrt{\frac{\sum_{i=1}^{n}(q_i^2 - \bar{q}_i^2)}{n}}} \times \frac{1}{s^{'}} $.
We choose to use $2^{-10}$ as $s^{'}$, which is the smallest power-of-two number that won't cause overflows in $q_i^{'}$.
Substituting $s^{'} = 2^{-10}$, the integer inference becomes:
\begin{eqnarray}
\bar{q} = round(\frac{\sum_{i = 1}^{n}(2^{10} q_i)}{n}) \\
\sigma = \sqrt{\frac{\sum_{i=1}^{n}(2^{20} q_i^2 - \bar{q}_i^2)}{n}} \\
q_i^{'} = round(\frac{(2^{10} q_i - \bar{q})}{\sigma}) \\
q_i^{''} = round(\frac{q_i^{'} q_{Wi} + q_{bi}}{2^{10}})
\end{eqnarray}
where $q_{Wi}$ is the quantized value of weight and $q_{bi}$ is quantized value
of bias. Weights are quantized with scale $s_{W} = \frac{range}{127}$ and the
scale of bias $s_b$ is the product between $2^{-10}$ and $s_{W}$.

\section{Experiments} \label{Experiments}
\subsection{Model Architecture Details: RNN-Transducer}\label{modelarch}
\begin{figure}
  \centering
  \includegraphics[width=0.7\columnwidth]{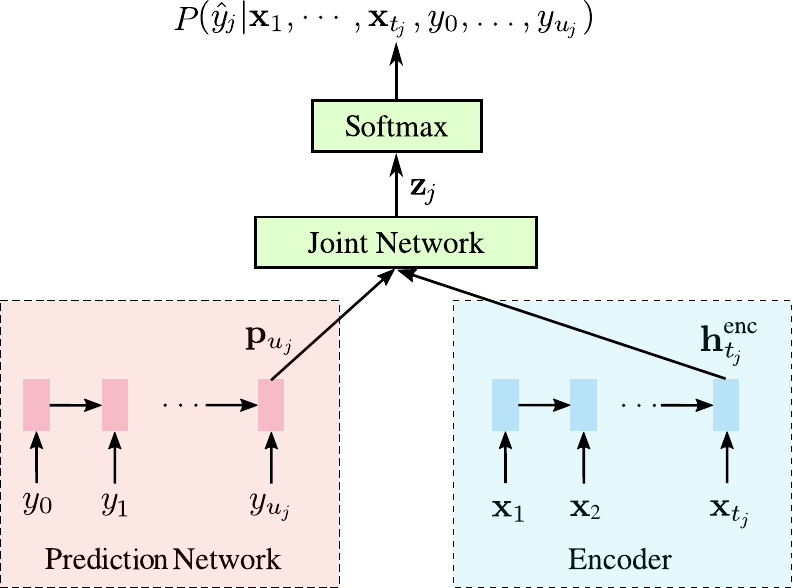}
  \caption{A schematic representation of CTC and RNNT, from~\cite{rohitpaper}.
  The \emph{encoder} takes input acoustic frames and plays the analogous
  role of a traditional acoustic model. The \emph{prediction
  network} (a.k.a. decoder) functions as a kind of language model.
  The \emph{joint} network combines the outputs of the previous two parts
  and leads finally to the softmax output.}
  \label{fig:ctcrnnt-schematic}
\vspace{-0.1in}
\end{figure}

The speech recognition architecture at the core of our experiments is
the RNN Transducer (RNN-T)~\cite{Graves12,graves2013speech}, depicted in
Figure~\ref{fig:ctcrnnt-schematic}.

Our base model is an RNN-T with a target word-piece output size of 4096,
similar to the model proposed by He et al.~\cite{he2019streaming}.
The encoder network consists of 8 LSTM layers and the prediction network
2 layers. Each layer consists of 2048 hidden units followed by a projection
size of 640 units. The joint network has 640 hidden units and a softmax layer with 4096 units.
Our baseline LSTM RNN-T model has 122M parameters in total.

In training, the learning rate ramps up to 1e-3 linearly in the first 32k
steps and decays to 1e-5 exponentially from 100k to 200k steps. We evaluate
all models on 3 test datasets with utterances from same domain as used in
training: the VoiceSearch, the YouTube and the Telephony dataset.
VoiceSearch and Telephony have average utterance length of 4.7s.
YouTube contains longer utterances, averaging 16.5 min per utterance.

\subsection{Pruning}
\begin{table*}[ht!]
\centering
\begin{tabular}{|l|l|c|l|l|} \hline
  \multirow{2}{*}{Sparsity} & \#Params (millions)  & \multicolumn{3}{c|}{WER}  \\\cline{2-5}
                             & \% of baseline & VoiceSearch  & YouTube & Telephony \\\hline
   0\%                       & 122.1 (100\%)  & 6.6          & 19.5    & 8.1  \\\hline
   50\%                      & 69.7 (57\%)    & 6.7          & 20.3    & 8.2  \\\hline
   70\%                      & 48.7 (39.9\%)  & 7.1          & 20.6    & 8.5  \\\hline
   80\%                      & 38.2 (31.3\%)  & 7.4          & 21.2    & 8.9  \\\hline
\end{tabular}
\caption{Comparison of pruned models with different sparsities}\label{tab:pruning_exp}
\end{table*}

We first pre-train a base model as described in Section \ref{modelarch}.
Then we apply the pruning
algorithm mentioned in Section \ref{pruning} to the weight matrices
in each LSTM layer of the RNNT model. The sparsity increases polynomially from 0 to
target sparsity within the first 100k.
In order to speed up inference on modern CPUs, a $16 \times 1$ block
sparse structure is enforced on cell.
Table~\ref{tab:pruning_exp} shows the
Word Error Rates (WER) and number of params of base model and pruned models at
different sparsity levels.

\subsection{Comparing RNN Topologies}
From our experiments, we learnt that that CIFG based RNN-T models are comparable to
LSTM-based RNN-T models in its performance. We experimented with using SRU
in the encoder and decoder layers and learnt that SRU layers could effectively
substitute LSTM layers in the decoder, but did not perform comparable to LSTM
layers in the encoders.

Table~\ref{tab:tab3} shows our finalized RNN-T model with sparse CIFG
(50\% weight sparsity) layers in the encoders,
and sparse SRU (30\% weight sparsity) layers in the decoder. This model has 59\%
fewer parameters than the baseline LSTM-based model, but only degraded by
7.5\% and 1.2\% of WER on VoiceSearch and Telephony test sets. Its WER on
YouTube has improved by 3.1\%. We show the results of a dense LSTM model (labeled ``Small") with the number of
hidden layer cells and projection layer cells at $0.7\times$ the those of the
baseline model. This smaller model is 45\% smaller than the original model but
suffers a WER degradation of 18.2\% on VoiceSearch - much worse than the sparse
models.

\begin{table*}[ht!]
\centering
\begin{tabular}{|l|l|l|c|l|l|l|l|l|}\hline
                         & Enc \& Dec Cell   & Sparsity  & \#Params\/M& Size      & \multicolumn{3}{c|}{WER}           & RT(0.9)          \\\hline
                         &                   &           & \% baseline   & MB        & VoiceSearch  & YouTube & Telephony &  \\\hline 
LSTM $0.7 \times |cell|$ & LSTMx8           & -         & 67.7          & 65      & 7.8          & 20.3    & 8.6       & 0.681       \\
(Small)                  & LSTMx2            & -         & (55.4\%)      &         &              &         &           &             \\\hline
CIFG-SRU                 & CIFG-LSTMx8       & -         & 89.6          & 86      & 6.9          & 19.1    & 7.8       & 0.806       \\   
                         & SRUx2             & -         & (73\%)        &         &              &         &           &             \\\hline
sparse-CIFG              & CIFG-LSTMx8       & 50\%      & 50.6          & 55      & 7.1          & 18.9    & 8.2       & 0.704       \\  
sparse-SRU               & SRUx2             & 30\%      & (41\%)        &         &              &         &           &             \\\hline
\end{tabular}
\caption{Comparing smaller dense LSTM model with models trained with sparse
CIFG-LSTM and SRU cells. RT factor is calculated using hybrid quantization on
Pixel3 small cores. Models are quantized using hybrid quantization.}\label{tab:tab3}
\end{table*}

To summarize, although SRU layers are less expressive to model long-term dependencies between the
RNN-T encoders cells, they were smaller and effective substitutes of LSTM
layers in the RNN-T decoder. A combination of 50\% sparse CIFG (encoder layers)
and 30\% sparse SRU (decoder layers) eliminated 59\% of the parameters with
respect to the baseline RNN-T model with a small loss of WER.

\subsection{Quantized LSTM}
\begin{table*}[ht!]
\centering
\begin{tabular}{|l|l|l|l|l|c|l|l|l|} \hline
              & Enc\&Dec & Sparsity  &\#Params(M)  & Quantization            & \multicolumn{3}{c|}{WER}           & RT(0.9) \\\hline
              &          &           &\% baseline  & Size(MB)                & VoiceSearch  & YouTube & Telephony &          \\\hline
LSTM          & LSTMx8   & 0\%       & 122.1       & Float,466MB           & 6.6          & 19.5    & 8.1       & 3.223    \\
(baseline)    & LSTMx2   & 0\%       & 100\%       & Hybrid,117MB           & 6.7          & 19.8    & 8.2       & 1.024    \\
              &          &           &             & Integer,117MB          & 6.7          & 19.8    & 8.2       & 1.013    \\
\hline
Sparse LSTM   & LSTMx8   & 50\%      & 69.7        & Float,270MB            & 6.7          & 20.2    & 8.2       & 1.771   \\
              & LSTMx2   & 50\%      & 57\%        & Hybrid,71MB            & 6.8          & 20.4    & 8.4      & 0.888     \\
              &          &           &             & Integer, 71MB           & 6.9          & 22.9    & 8.7    & 0.869     \\
\hline
Sparse CIFG   & CIFGx8   & 50\%      & 56.3        & Float,219MB            & 7.1          & 21.7    & 8.3   & 1.503   \\
              & CIFGx2   & 50\%      & 46\%        & Hybrid,57MB            & 7.2          & 21.4    & 8.5   & 0.743     \\
              &          &           &             & Integer,57MB           & 7.2          & 20.6    & 8.7    & 0.709     \\
\hline
\end{tabular}
\caption{Comparison of float, hybrid and fully quantized models. RT factor is calculated on
Pixel3 small cores.}
\label{tab:benchmarktab}
\end{table*}

The accuracy and CPU performance comparison between float, hybrid and fully
quantized models is listed in Table~\ref{tab:benchmarktab}.
The results show that our proposed integer quantization
has negligible accuracy loss on Voice Search. Longer utterances
(in YouTube), which are typically more challenging
for sparse models, still see comparable results from sparse models as from float models.

We use Real Time (`RT') factor, which is the ratio between the wall time needed
for completing the speech recognition and the length of the audio, to measure the
end-to-end performance. We denote RT(0.9) as the RT factor at 90 percentile. Using
the \cite{tflite} runtime in a typical mobile CPU (Pixel 3 small cores),
we compare the RT(0.9) between the float, hybrid and integer models, and see the integer model achieve
an RT factor that is around $30\%$ with respect to the float model.

\section{Conclusion}\label{conclusions}
In this work we present a comprehensive set of optimizations spanning from
more efficient neural network building blocks to the elimination, and reduction
in precision, of neural network parameters and computations. Altogether they
result in a high quality speech recognition system ``Sparse CIFG, Sparse SRU"
that is 8.5x smaller in size (466MB to 55MB), and 4.6x faster in RT factor
(3.223 to 0.704) when comparing to our full precision ``LSTM Baseline".
We validate that neural connection pruning is a useful
tool to shrink neural network at the cost of a relatively small accuracy loss.
Moreover, the use of a particular ``block
sparsity" configuration enables execution speedups in CPUs widely used
in mobile, desktop and server devices, without requiring specialized hardware
support. We also demonstrate that RNN variants result in competitive qualitative
performance with respect to the widely accepted and used LSTM topology,
while also reducing the number of parameters and potentially enabling other
optimizations. CIFG-LSTM, in particular, is an underused simple optimization
to take advantage of. We show that quantization is an
effective technique that makes neural network models smaller and faster when
inferencing in CPUs. Integer quantization opens the door towards using more specialized
neural network acceleration chips such as Tensor Processing Units. Finally, we
verify that all these techniques are complimentary to each other. Although
the accuracy losses of each technique do compound, they do not do it in a way that multiply
each other with catastrophic results. On the contrary, our smallest model
"Sparse CIFG" achieves better accuracy, even quantized, than that of a
small baseline model ``LSTM (Baseline small)" evaluated in full precision.

\bibliography{main}
\bibliographystyle{sysml2019}

\end{document}